\def\year{2018}\relax
\documentclass[letterpaper]{article} 
\usepackage{aaai}  
\usepackage{times}  
\usepackage{helvet}  
\usepackage{courier}  
\usepackage{url}  
\usepackage{graphicx}  
\usepackage[utf8]{inputenc} 
\usepackage[T1]{fontenc}    
\usepackage{hyperref}       
\usepackage{url}            
\usepackage{booktabs}       
\usepackage{amsfonts}       
\usepackage{nicefrac}       
\usepackage{microtype}      
\usepackage{placeins} 
\usepackage{color}
\usepackage{subcaption}
\usepackage{enumitem}
\usepackage{mathtools}
\usepackage{wrapfig}
\usepackage{amsmath}
\usepackage{amssymb}
\usepackage[toc,page]{appendix}
\usepackage{pifont}

\newcommand{\dataset}{TallyQA}
\newcommand{\inline}[1]{\citeauthor{#1}~\shortcite{#1}}
\def\eg{\emph{e.g}. } 
\def\ie{\emph{i.e}. } 

\def\etal{\emph{et al}. }

\begin{document}
%

\title{TallyQA: Answering Complex Counting Questions}
\author{Paper ID: 595}
\author{Manoj Acharya, Kushal Kafle, Christopher Kanan\\
Chester F. Carlson Center for Imaging Science\\Rochester Institute of Technology \\ 
{\tt\small \{ma7583, kk6055, kanan\}@rit.edu}}
\nocopyright
\maketitle
\begin{abstract}
Most counting questions in visual question answering (VQA) datasets are simple and require no more than object detection. Here, we study algorithms for complex counting questions that involve relationships between objects, attribute identification, reasoning, and more. To do this, we created TallyQA, the world's largest dataset for open-ended counting. We propose a new algorithm for counting that uses relation networks with region proposals. Our method lets relation networks be efficiently used with high-resolution imagery.  It yields state-of-the-art results compared to baseline and recent systems on both TallyQA and the HowMany-QA benchmark. 
\end{abstract}

\section{Introduction}
\begin{figure}[t!]
  \centering
  \includegraphics[width=.43\textwidth]	{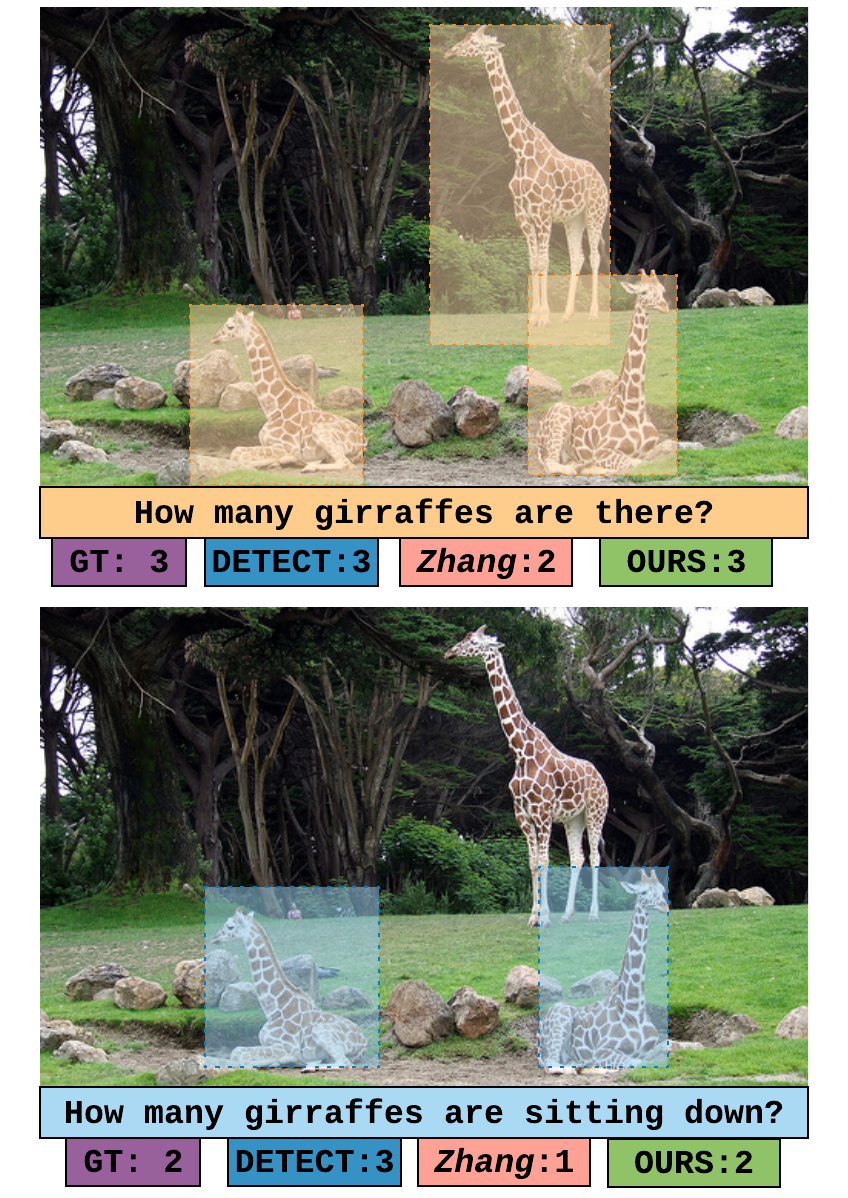} \\
        \label{fig:complex-example}
\caption{Counting datasets consist mostly of simple questions (top) that can be answered solely using object detection. We study complex counting questions (bottom) that require more than object detection using our new TallyQA dataset. }
\label{fig:main}
\end{figure}

Open-ended counting systems take in a counting question and an image to predict a whole number that answers the question. While object recognition systems now rival humans~\cite{he2016deep}, today's best open-ended counting systems perform poorly~\cite{kafle2017analysis,chattopadhyay2016counting}. This could be due to an inability to detect the correct objects or due to an inability to reason about them. To address this, we distinguish between \emph{simple} and \emph{complex} counting questions (see Fig.~\ref{fig:main}). Simple counting questions only require object detection, e.g., ``How many dogs are there?'' Complex questions require deeper analysis, e.g., ``How many dogs are eating?''

Open-ended counting is a special case of visual question answering (VQA)~\cite{antol2015vqa,malinowski2014multi}, in which the goal is to answer open-ended questions about images. The best VQA systems pose it as a classification problem where the answer is predicted from convolutional visual features and the question~\cite{kafle2017-cviu}. While this succeeds for many question types, it works poorly for counting~\cite{kafle2017analysis,chattopadhyay2016counting}. Recently, better results were achieved by using region proposals generated by object detection algorithms~\cite{trott2017interpretable,zhang2018learning}. However, datasets mostly contain simple counting questions, as shown in Table~\ref{table:dataset-count-summary}. Due to their rarity, complex questions need to be analyzed separately to determine if a model is capable of answering them. 

\textbf{This paper makes three major contributions:} 
\begin{enumerate}[nolistsep,noitemsep]
\item We describe TallyQA, the world's largest open-ended counting dataset, which is over 2.5 times bigger than VQA2. TallyQA is designed to study both simple questions that require only object detection and complex questions that demand more. It will be made publicly available.

\item We propose the relational counting network (RCN), a new algorithm for counting that infers relationships between objects and background image regions. It is inspired by relation networks, with modifications to handle a dynamic number of image regions and to explicitly incorporate background information. 

\item We show that RCN surpasses state-of-the-art methods for open-ended counting on both TallyQA and the HowMany-QA benchmark.

\end{enumerate}
\section{Related Work}
\subsection{VQA Datasets \& Counting}
Popular VQA datasets contain a significant number of counting questions, e.g., about 7\% in COCO-QA~\cite{ren2015exploring}, 10\% of VQA1~\cite{antol2015vqa}, 10\% of VQA2~\cite{goyal2016making}, and 10\% of TDIUC~\cite{kafle2017analysis}. There are also counting specific VQA datasets. CountQA~\cite{chattopadhyay2016counting} was created by importing question-answer (QA) pairs from the validation split of VQA1 and COCO-QA. This small dataset has only 2,287 test QA pairs. Recently, HowMany-QA~\cite{trott2017interpretable} was created by importing QA pairs from Visual Genome and VQA2, and it is considered the gold standard for open-ended counting. As shown in Table~\ref{table:dataset-count-summary}, complex questions are scarce in these datasets. Simple questions can be solved using solely an object detection algorithm, so they do not appropriately test a system's ability to answer arbitrary counting questions, including those requiring reasoning or attribute recognition. Our new dataset, TallyQA, is designed to evaluate both simple and complex counting questions, enabling these and other capabilities to be appropriately evaluated. Note that we have limited our discussion to natural language VQA systems which pose different challenges compared to VQA on synthetic datasets, which are often designed for specific purposes \cite{clevr,kafle2018dvqa,zhang2016yin}.

\subsection{Algorithms for Open-Ended Counting}
Open-ended counting systems take as input a ``How many ...?'' question and an image and then output a count. This is a VQA sub-problem. For counting, there are two general approaches. The first involves inferring the count directly in an end-to-end framework operating on high-level CNN features. The second approach is to detect object bounding boxes or region proposals, and then aggregate question relevant bounding boxes. While there are many direct methods, over the past year region-based schemes for open-ended counting have been studied.

\subsubsection{Direct Methods.}

State-of-the-art VQA systems train a classifier to predict the answer from the image and question. Typically, image features are encoded using a CNN that was pre-trained on ImageNet, and questions are encoded using a recurrent neural network (RNN). Many innovations involve different ways of combining image and question features~\cite{lu2016hierarchical,fukui2016multimodal,ben2017mutan,kafle2016answer}, modular networks \cite{andreas2016learning}, data-augmentation \cite{kafle2017data} among many others.

Direct methods perform poorly at counting for \emph{real-world} image datasets. In \inline{kafle2017analysis}, three state-of-the-art VQA algorithms were compared to baselines on TDIUC's counting questions. The best performing method, MCB, achieved  51\% accuracy, which was only 6\% better than an image-blind (question-only) model. This was true even though most of TDIUC's counting questions are simple. This suggests these methods are primarily exploiting scene and language priors. For VQA2, the best method~\cite{teney2017tips} of the CVPR-2017 VQA Workshop challenge achieved 69\% overall, but only  47\% accuracy on number questions, most of which are counting.

We hypothesize that the inability of VQA algorithms to count is due to the way their architectures are designed. These systems operate on image embeddings computed using a CNN. Mean pooling and weighted mean pooling (attention) operations may destroy information that can be used to determine how many objects of a particular type are present. 

\begin{table}[t]
\renewcommand{\tabcolsep}{2pt}
\begin{tabular}{rrrrr}
\toprule
&  VQA2 & TDIUC & HowMany-QA & \textbf{\dataset~(Us)} \\
\midrule
Simple  &         78,455  & 148,719  &  68,956  & 211,430 \\
Complex &         34,799  & 16,043    &  37,400 & 76,477 \\ \midrule
\textbf{Total}&   113,254 & 164,762   &  106,356 & 287,907 \\
\bottomrule
\end{tabular}
\caption{The number of counting questions for previous VQA datasets compared to TallyQA dataset. \label{table:dataset-count-summary}  }
\end{table}

\subsubsection{Counting Specific Systems.}

While counting has long been studied for specific computer vision problems~\cite{zhang2015cross,dalal2005histograms,wang2011automatic,ryan2009crowd,ren2016end}, only recently has open-ended counting in natural scenes been studied. \inline{chattopadhyay2016counting} studied open-ended counting in typical scenes, and they evaluated three counting-specific methods: DETECT, GLANCE, and SUBITIZE. DETECT is built on top of an object detection algorithm, which was Fast R-CNN~\cite{girshick2015fast} in their implementation. DETECT works by finding the first noun in a question and then matching it to the closest category the detection algorithm has been trained for (e.g., COCO objects). GLANCE uses a shallow multi-layer perceptron (MLP) to regress for specific object counts from a CNN embedding, with the appropriate output unit chosen based on the first noun. SUBITIZE involves breaking the image into a grid, extracting image embeddings from each grid location,  aggregating information across grids using an RNN, and then predicting the count for every class in the dataset. Although none of these methods are capable of handling complex questions, all of them outperformed MCB, which was a state-of-the-art VQA model. 
\begin{figure*}[h]
  \centering
    \begin{subfigure}[b]{0.32\textwidth}
    	\centering        
        \includegraphics[width=\textwidth]{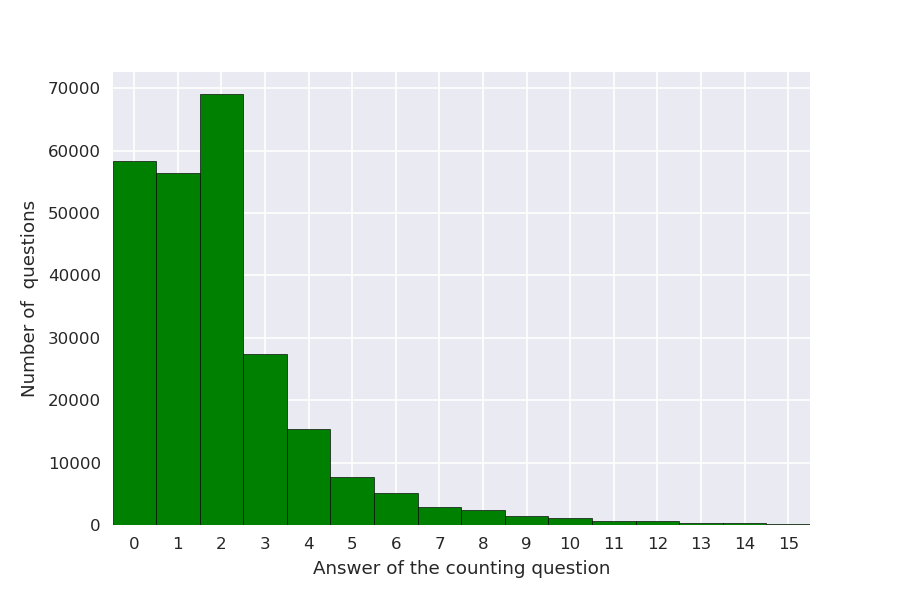}
        \caption{Train}
        \label{fig:train-hist}
    \end{subfigure} ~
     \begin{subfigure}[b]{0.32\textwidth}
        \centering        
        \includegraphics[width=\textwidth]{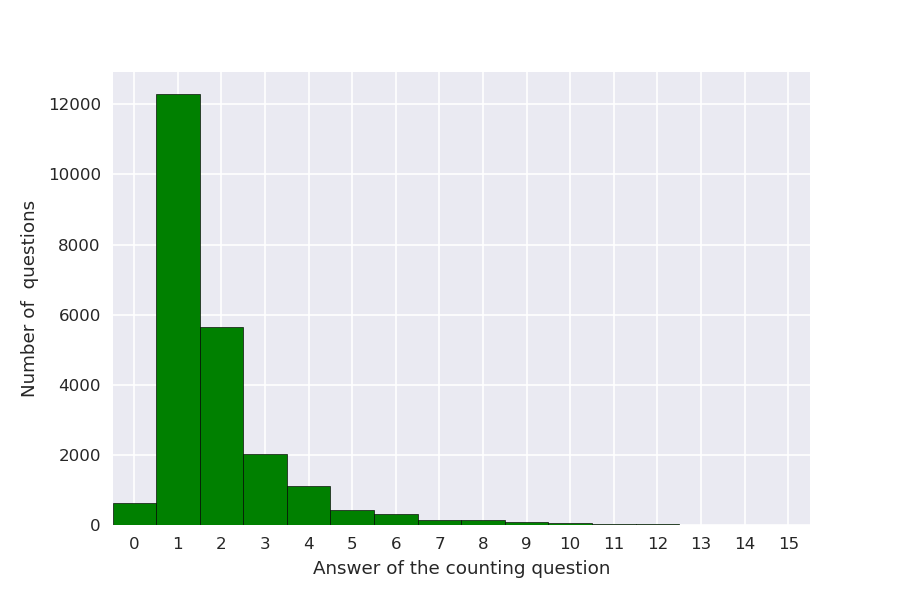}
        \caption{Test-Simple}
        \label{fig:test-simple-hist}
    \end{subfigure}~
     \begin{subfigure}[b]{0.32\textwidth}
        \centering        
        \includegraphics[width=\textwidth]{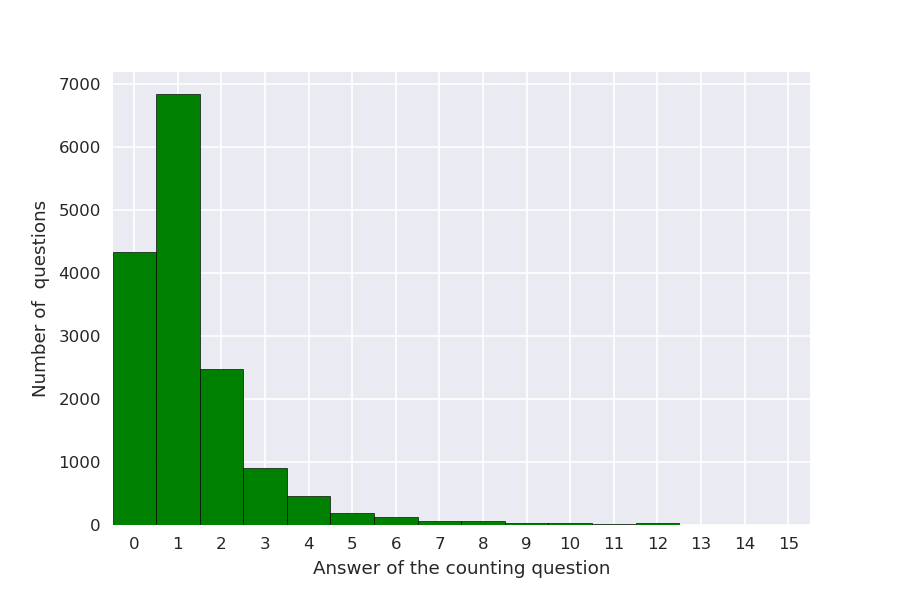}
        \caption{Test-Complex}
        \label{fig:test-simple-hist}
    \end{subfigure}~
\caption{Histogram of answer counts for each of the three splits of \dataset.\label{fig:count-frequency}}
\label{fig:distribution}
\end{figure*}

Recently, \inline{trott2017interpretable} and \inline{zhang2018learning} both created algorithms for open-ended counting in natural scenes that are built on top of object proposals generated by an object detection algorithm trained on Visual Genome.  Trott~\etal created the ILRC algorithm, which redefines counting as a sequential object selection problem.  ILRC uses reinforcement learning to select the objects that need to be counted based on the question. Zhang~\etal created a method that uses object detection and then constructs a graph of all detected objects based on how they overlap. Edges in the graph are removed based on several heuristics to ensure that duplicated objects are only counted once.

Both Trott~\etal and Zhang~\etal operate on region proposals and loosely based on the idea of filtering out irrelevant boxes based on the question, \ie selecting a subset of question relevant region proposals. However, successfully determining which boxes should be counted for a given question often requires comparing it with other object proposals (required for duplicate detection, comparative and positional reasoning, etc.), and the background (for modeling context, finding relative size, etc.). Since neither of these algorithms performs any relational or comparative reasoning between the boxes, they may have an impaired ability to answer complex questions. Here, our RCN model applies relational reasoning to object-object and object-background pairs, giving it a more robust capability to answer complex and relational questions. Indeed, our experiments show that RCN outperforms other models on complex questions.

\section{The \dataset~Dataset}
Complex counting questions are rare in existing datasets. This prompted us to create TallyQA. TallyQA's test set is split into two parts: Test-Simple for simple counting questions and Test-Complex for complex counting questions. We gathered new complex questions using Amazon Mechanical Turk (AMT), and imported both simple and complex questions from other datasets. Table~\ref{table:dataset-count-summary} shows the total number of questions in \dataset~ compared to others, and it has over twice as many complex questions. The number of questions in the train and test sets by  source is given in Table~\ref{tab:dataset-simple-complex-counts}.  Fig.~\ref{fig:result-images} shows example images and questions. Test-Simple and Test-Complex contain images from only Visual Genome, and Train has images from COCO and Visual Genome.

\subsection{Collecting New Complex Questions}

To gather new complex questions, we developed targeted AMT tasks that yielded 19,500 complex questions for 17,545 unique images. These tasks were designed to fight the biases in earlier datasets, where simple counting questions were predominantly asked~\cite{kafle2017-cviu}. TallyQA's images are drawn from both COCO and Visual Genome, which provides more variety than COCO alone. About 800 unique annotators provided QA pairs. For all tasks, annotators were not allowed to submit obviously simple questions, e.g., ``How many $x$?'' and ``How many $x$ in the photo?''  We manually checked AMT questions to ensure they were complex, and we removed poor quality questions.  
\begin{table}[t]
\centering
\label{table:dataset}
\begin{tabular}{lrr}
\toprule
\textbf{Split} & \textbf{Questions} & \textbf{Images} \\ \midrule
Train & 249,318 & 132,981 \\    
\quad\textit{AMT} & 3,902 & 3,494\\
\quad\textit{Imported} &245,416 &129,487  \\
Test-Simple & 22,991 & 18,411 \\
\quad\textit{AMT} & 0 & 0 \\
\quad\textit{Imported} & 22,991 & 18,411 \\
Test-Complex & 15,598 & 14,051 \\
\quad\textit{AMT} & 15,598 & 14,051 \\
\quad\textit{Imported} & 0 & 0 \\
\bottomrule 
\end{tabular}
\caption{Number of questions and images in \dataset. \label{tab:dataset-simple-complex-counts}}
\end{table}

We endeavored to ensure non-zero complex questions were \emph{difficult}, e.g., ``How many men are wearing glasses?'' is not difficult if all of the men in the image are wearing glasses. To do this, annotators were told to ask questions in which there were counter examples, e.g., to ask ``How many men are wearing glasses?'' only if it had an answer greater than zero, and the contrary question ``How many men are \emph{not} wearing glasses?'' had an answer greater than zero.

We created a separate task to generate hard complex questions with zero as the answer. Annotators were asked to make questions about objects with attributes not observed in the image, e.g., asking the question ``How many dogs have spots?'' when there was a dog \emph{without} spots in the image. Similar examples were shown to annotators before annotation. 

\subsection{Importing Questions from Other Datasets}

\dataset~also contains questions imported from VQA2 and Visual Genome. A similar approach was used to create HowMany-QA~\cite{trott2017interpretable} and TDIUC~ \cite{kafle2017analysis}. We imported all questions beginning with the phrase ``How many...'' with answers that were whole numbers between 0--15. Following \inline{kafle2017analysis}, for VQA2, we required that 5 of the 10 annotators give the same answer. Although these questions were generated by humans, as seen in Table~\ref{table:dataset-count-summary}, most are simple.

We also imported synthetic counting questions from TDIUC~\cite{kafle2017analysis}. These questions were generated for COCO images using its semantic annotations. The creators used a variety of templates to introduce variation in the questions and used heuristics to avoid answer ambiguity. All template generated questions from TDIUC are simple. In addition to templates, we used their method for making ``absurd'' questions to create both simple and complex zero count questions. To do this, we first find the objects absent from an image based on its COCO annotations. Then, we randomly sample the counting questions from the rest of the dataset that ask about counting these objects.

\subsection{Classifying Simple and Complex Questions}

The Test-Complex dataset was made using only new, human vetted complex questions from AMT. Because simple questions are common in existing datasets like VQA2, we used imported questions to make Test-Simple. To do this, we developed a classifier to determine if a question was simple. 

Our simple-complex classifier is made from a set of linguistic rules. First, any substrings such as ``...in the photo?'' or ``...in the image?'' were removed from the question. Then, we used SpaCy to do part of speech tagging on the remaining substring. It was classified as simple if it had only one noun, no adverbs, and no adjectives, otherwise it was deemed complex. This will classify questions such as ``How many dogs?'' as simple and ``How many brown dogs?'' as complex. 

Every question classified as simple by our rules will be correct (i.e., the false positive rate is zero), making it suitable for creating Test-Simple, but it may sometimes classify simple questions as complex (i.e., the false negative rate is non-zero). For example, the question ``How many men are wearing red hats to the left of the tree?'' would be classified as complex by our classifier. However, if there was only a single person in the image then it is not truly a complex question, despite the apparent complexity. These kinds of questions are rare and our simple-complex classifier works robustly, but it is possible that it will underestimate the number of simple questions and overestimate the number of complex when used to characterize a dataset. For this reason, we only use  human-vetted questions in TallyQA's Test-Complex set.
\subsection{Dataset Splits \& Statistics}

TallyQA is split into one training split (Train) and two test splits: Test-Simple and Test-Complex. Using our simple-complex classifier, Train was found to have 188,439 simple and 60,879 complex questions. The number of questions in each split is given in Table~\ref{tab:dataset-simple-complex-counts}. The test splits are comprised exclusively of Visual Genome imagery, and no images in the test splits are used in training.

\section{A New Framework for Complex Counting}
\label{section:framework}

\begin{figure*}[t]
\centering
\includegraphics[width=1\textwidth]{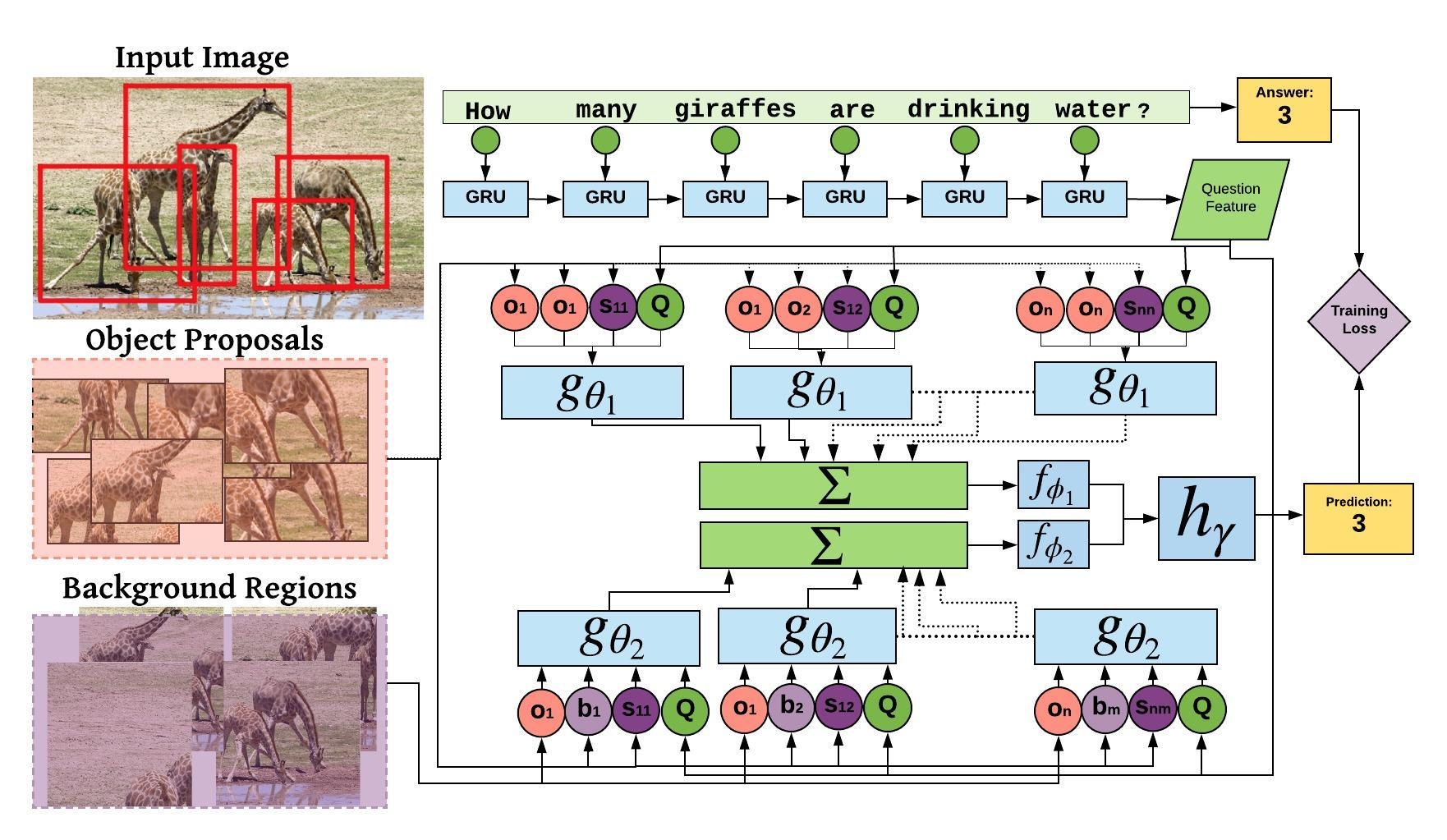}
\caption{Our RCN model computes the relationship between foreground regions as well as the relationships between the these regions and the background to efficiently answer complex counting questions. In this example, the system needs to look at the relationship of each giraffe to each other and with the water (background).}
\label{fig:system}
\end{figure*}

Our RCN model, depicted in Fig.~\ref{fig:system}, is formulated as a modified relation network (RN)~\cite{santoro2017simple} that can reason about the nature of relationships between image regions. RCN uses the question $Q$ to guide its processing of a list of $n$ foreground region proposals, $O = \{o_1, o_2, \ldots , o_n \}$, and $m$ background regions, $B = \{b_1, b_2, \ldots , b_m \}$, with $o_i \in \mathbb{R}^K$ and $b_j \in \mathbb{R}^K$.  Formally, our RCN model is the combination of two RN sub-networks, i.e.,
\begin{align} 
    \text{Count}(O,B,Q) = h_{\gamma}\left( \text{RN}(O, O) \oplus \text{RN}(O, B) \right),
      \label{eq:RNall}
\end{align} 
where $\oplus$ denotes concatenation,  $RN(O,O)$ represents the RN that infers the relationship between foreground regions, $RN(O,B)$ represents the RN responsible for inferring the relationship between each foreground and background region, and $h_{\gamma}$ is a neural network with parameters $\gamma$ that predicts the final count.

The RN for predicting the relationship between foreground proposals in the context of question $Q$ is given by
\begin{align} 
    \text{RN}(O,O) = f_{\phi_1}\left(\sum_{i,j}g_{\theta_1}(o_i, o_j, s_{ij},Q)\right),
      \label{eq:RN}
\end{align} 
where $f_{\phi_1}$ and $g_{\theta_1}$ are neural networks with parameters $\phi_1$ and  $\theta_1$, respectively, that each  output a vector, and the vector $s_{ij}$ encodes spatial information about the $i$-th and $j$-th proposals. Like the original RN model, the sum is computed over all $n^2$ pairwise  combinations.   Similarly, the RN for predicting the relationship of each proposal to the background is given by,
\begin{align} 
    \text{RN}(O,B) = f_{\phi_2}\left(\sum_{i,j}g_{\theta_2}(o_i, b_j, s_{ij}, Q)\right),
      \label{eq:RNob}
\end{align} 
where  $f_{\phi_2}$ and $g_{\theta_2}$ are neural networks with parameters $\phi_2$ and  $\theta_2$, respectively, that  output vectors.
RCN has two major innovations over the original RN approach. 

The original RN used raw CNN feature map indices as regions. This worked well for CLEVR, but this approach works poorly for real-world VQA datasets that require processing at higher resolutions (e.g., VQA2).  RCN overcomes this problem by using region proposals. As input, the original RN model used the $d^2$ elements in a $d \times d$ convolutional feature map, which were each tagged with their spatial coordinates. This means it computed $d^4$ pairwise relationships. For recent direct VQA methods, a CNN feature map is typically $14 \times 14$, meaning that 38,416 comparisons would be needed per counting query. In contrast, RCN's proposal generator produces only 31.12 foreground regions and 16 background patches per image, so only $31.12^2 + (31.12 \times 16) = 1466$ comparisons are made, on average. By using proposals, RCN reduces the number of comparisons by a factor of 26 and scales to real-world imagery, whereas the original RN model used lower resolution imagery and was only evaluated on CLEVR~\cite{clevr}, a synthetic dataset that has simple geometric shapes and a plain background.

RCN's second innovation is the explicit incorporation of the background. For queries such as ``How many dogs are laying in the grass?'' it is necessary to consider background entities (stuff) that are ignored by object detection systems.  RCN uses $m$ image background patches, and computes the relationships of each region with each background patch, enabling the background to be studied with relatively few comparisons. In contrast, the original RN model did not explicitly deal with the background, but it was likely unnecessary due to the simple scenes in CLEVR.  Explicitly modeling the background can help  answer complex counting questions, which often involve attributes of background objects or relationships between objects and background entities.

Internally, RCN uses the spatial relationship between regions $o_i$ and $o_j$ to help predict the count. Using $s_{ij}$ is critical to ensuring each object is counted only once during prediction, and it enables RCN to learn to do non-maximal suppression to cope with overlapping proposals. The spatial relationship vector is given by 
\begin{align} 
    \text{$s_{ij}$} = \left[ \ell_i , \ell_j , \xi_{ij},  IoU_{ij} , \frac{IoU_{ij} }{ A_{i}} , \frac{IoU_{ij} }{ A_{j}} \right],
      \label{eq:spatial}
\end{align} 
where $\ell_i$ and $\ell_j$ encode the spatial information of each proposal individually, $A_i$ and $A_j$ are the area of proposals, $\xi_{ij}$ is the dot product between each proposal's CNN features to model how visually similar they are, and $IoU_{ij}$ is the intersection over union between the two proposals. The vector $\ell_i = \left[\frac{x_{min}}{W},\frac{y_{min}}{H} , \frac{x_{max}}{W},\frac{y_{max}}{H}, \frac{x_{max}- x_{min}}{W} , \frac{y_{max} - y_{min}}{H}\right]$,  where $(x_{min},x_{max})$ and $(y_{min},y_{max})$ represent the top-left and bottom-right corners of proposal $i$, and $W$ and $H$ are the width and height of the image, respectively. 

\subsection{Training \& Implementation Details}
The question $Q$ is embedded using a one layer GRU that takes as input pre-trained 300-dimensional Glove vectors for each word in the question~\cite{pennington2014glove} and outputs a vector of 1024 dimension. The GRU used in all models are regularized using a dropout of 0.3.

For foreground proposals, we use the boxes and CNN features produced by Faster R-CNN~\cite{ren2015faster} with ResNet-101 as its backbone. The Faster R-CNN model is trained to predict boxes in Visual Genome, which contain a wide variety of objects and attributes. This approach for generating proposals was pioneered by \inline{Anderson2017up-down}, and has since been used by multiple VQA systems.

For the background patches, we extract ResNet-152 features from the entire image before the last pooling layer and then apply average pooling over these features to reduce them to a $4 \times 4$ grid. Each of these 2048-dimensional vectors represents a $112 \times 112$ pixel background region. In RCN, $g_\theta$  has three hidden layers and $g_\phi$ has one hidden layer, which each consist of 1024 rectified linear units (ReLUs). The outputs of these networks are then concatenated and passed to $h_\gamma$, which has one hidden layer with 1024 units and ReLU activation. The softmax output layer treats counting as a classification task, and it is optimized using cross-entropy loss. RCN is trained using the Adam optimizer with a learning rate of $7e^{-4}$ and a batch size of 64 samples.

\section{Experiments}
\begin{table*}
\label{table1}
\centering
\footnotesize
\begin{tabular}{lrr|rr|rr}
 & \multicolumn{2}{c|}{HowMany-QA} & \multicolumn{2}{c|}{\dataset~Test-Simple} & \multicolumn{2}{c}{\dataset~Test-Complex} \\
 &ACC    & RMSE & ACC    & RMSE & ACC     & RMSE  \\ \midrule
Guess-1 & 33.8 & 3.74  & 53.5 &1.78  &43.9  &1.57  \\
Guess-2 & 32.1  & 3.34  & 24.5 &1.56 & 15.9  &1.69 \\
Q-Only &37.1  & 3.51 & 44.6 & 1.74  & 39.1  & 1.75 \\
I-Only &37.3 & 3.49  &46.1 & 1.71 & 26.4 & 1.69 \\
Q+I & 40.5 & 3.17  & 54.7 & 1.44 & 48.8 & 1.57  \\ 
DETECT &43.3 & 3.66   &50.6 &2.08  & 15.0  & 4.52 \\ \midrule 
MUTAN & 45.5 & 2.93 & 56.5 &1.51  & 49.1& 1.59   \\
Zhang~\etal & 54.7  & 2.59 & 70.5  & 1.15  & 50.9  & 1.58  \\
IRLC & 56.1 & 2.45  & -- & -- & -- & --   \\
RCN (Ours) & \textbf{60.3} & \textbf{2.35}  & \textbf{71.8} & \textbf{1.13} & \textbf{56.2} &\textbf{1.43} \\
\bottomrule
\end{tabular}
\caption{Performance breakdown on TallyQA and Howmany-QA datasets using accuracy (\%) and RMSE.  \label{tab:main-results}}
\end{table*}

In this section, we describe a series of experiments to evaluate the efficacy of multiple algorithms on both simple and complex counting questions.

\subsection{Models Evaluated}
We compare RCN against two state-of-the-art models specifically for open-ended counting: \inline{zhang2018learning} and IRLC~\cite{trott2017interpretable}. We also compare against MUTAN~\cite{ben2017mutan}, one of best direct VQA methods. Lastly, 
we compare RCN to six baseline counting models:
\begin{enumerate}[noitemsep,nolistsep]
\item \textbf{Guess-1}: Answer 1 for all questions.
\item \textbf{Guess-2}: Answer 2 for all questions.
\item \textbf{Q-Only}: An image-blind MLP model with a hidden layer of 1024 units that uses only the question. The question features are obtained from the last hidden layer of the same RNN architecture used by our RCN model.
\item \textbf{I-Only}: A question-blind  MLP  that  has one hidden layer with 1024 units.
\item \textbf{Q+I}: An MLP with 1024 hidden units that uses both image and question features. 
\item \textbf{DETECT}: DETECT is an upgraded version of the method from \cite{chattopadhyay2016counting}. The main difference is that we use the more recent YOLOv2~\cite{redmon2016yolo9000} method instead of Fast R-CNN. DETECT extracts the first noun from the question. It then finds the most semantically similar category that YOLOv2 was trained on to that noun based on word similarity, and then it outputs the total number of YOLOv2 boxes produced for that category. 
\end{enumerate}
MUTAN, I-Only, and Q+I use ResNet-152 features. Q-Only, I-Only, Q+I, MUTAN, Zhang~\etal, and RCN all use cross-entropy loss and treat counting as a classification problem. Before evaluation, the output of all models was rounded to the nearest whole number and constrained to be within the range of values in the datasets. 

\subsection{Results}

\begin{figure*}
\centering
\footnotesize
    \captionsetup[subfigure]{justification=centering}     
        \begin{subfigure}[t]{0.3\textwidth}
		\includegraphics[width=\textwidth, ]{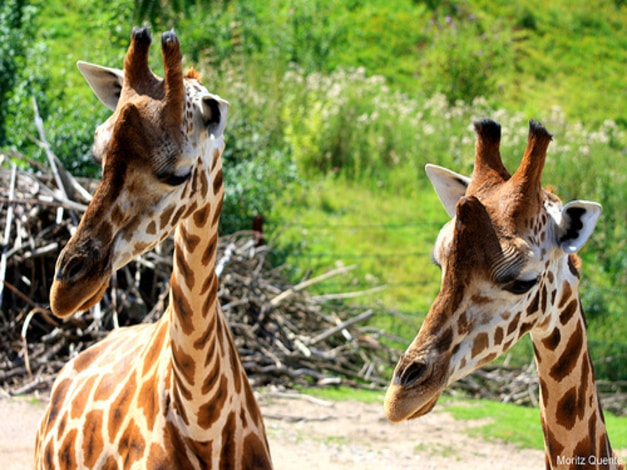}
        \caption{
	    How many giraffes are there? \\
\textcolor{green}{GT}: 2, \textcolor{red}{DETECT}: 2, \textcolor{blue}{Zhang}:2, \textcolor{magenta}{RCN}: 2 \label{subfig:d} }
    \end{subfigure}
    \hfill    
            \begin{subfigure}[t]{0.3\textwidth}
        	 	\includegraphics[width=\textwidth, ]{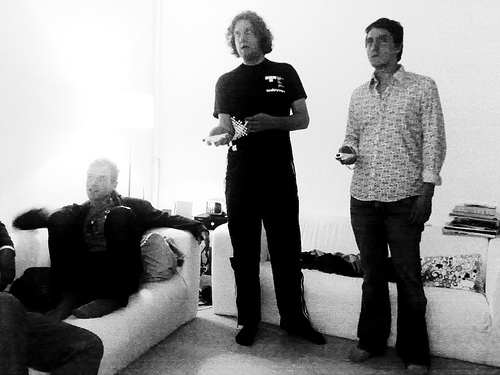}
        \caption{
	    How many people are standing?\\
        \textcolor{green}{GT}: 2, \textcolor{red}{DETECT}: 4, \textcolor{blue}{Zhang}: 3, \textcolor{magenta}{RCN}: 2 \label{subfig:e}}
    \end{subfigure}   
    \hfill
    \begin{subfigure}[t]{0.3\textwidth}
	 	\includegraphics[width=\textwidth, ]{imageseg/2348611.jpg}
\caption{How many people in the front row?\\
        \textcolor{green}{GT}: 8, \textcolor{red}{DETECT}: 22, \textcolor{blue}{Zhang}: 6, \textcolor{magenta}{RCN}: 8\\
		\label{subfig:c} }
\end{subfigure}
    \begin{subfigure}[t]{0.3\textwidth}
	 	\includegraphics[width=\textwidth, ]{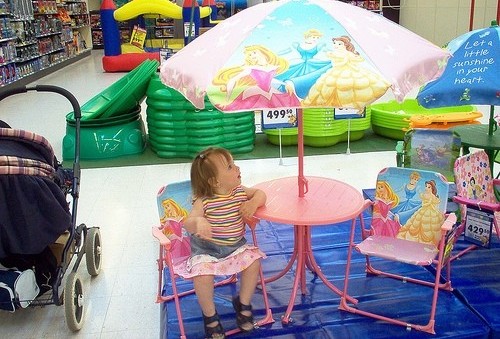}
        \caption{
	    How many chairs have a girl sitting on them?\\
        \textcolor{green}{GT}: 1, \textcolor{red}{DETECT}: 7, \textcolor{blue}{Zhang}: 2, \textcolor{magenta}{RCN}: 1\\ \label{subfig:b} }
	\end{subfigure}
    \hfill
    \begin{subfigure}[t]{0.3\textwidth}
		\includegraphics[width=\textwidth, ]{imageseg/2317119.jpg}
        \caption{
	    How many players are wearing red uniforms? \\
        \textcolor{green}{GT}: 3, \textcolor{red}{DETECT}: 11, \textcolor{blue}{Zhang}: 4, \textcolor{magenta}{RCN}: 3\\  \label{subfig:a}}   
	\end{subfigure}
    \hfill    
    \begin{subfigure}[t]{0.3\textwidth}
	 	\includegraphics[width=\textwidth, ]{imageseg/3999.jpg}
\caption{How many strings does the instrument to the left have?\\
       \textcolor{green}{GT}: 4, \textcolor{red}{DETECT}: 3, \textcolor{blue}{Zhang}: 1, \textcolor{magenta}{RCN}: 0\\
		\label{subfig:f} }
\end{subfigure}   
\caption{Example model outputs on \dataset. While other models fail at positional reasoning questions (\eg Fig.~\ref{subfig:c}), RCN can infer an object's relative position to other objects. \label{fig:result-images} Since RCN is based on region proposals, it struggles when proposals do not align with question relevant objects (Fig.~\ref{subfig:f}).}
    \end{figure*}

Results for all methods on HowMany-QA and both of \dataset's test sets are given in Table~\ref{tab:main-results}. Following earlier work~\cite{chattopadhyay2016counting,trott2017interpretable,zhang2018learning}, we compute both accuracy and RMSE. RMSE captures that larger errors should be penalized more heavily.

\begin{table}
\centering
\footnotesize
\begin{tabular}{lrr|rr}
  & \multicolumn{2}{c|}{Test-Simple} & \multicolumn{2}{c}{Test-Complex} \\
 & ACC    & RMSE & ACC     & RMSE  \\ \midrule
RCN  -- No Background  & 69.4 & 1.18 & 51.8 & 1.50 \\
RCN -- Full  & \textbf{71.8} & \textbf{1.13} & \textbf{56.2} & \textbf{1.43} \\
\bottomrule
\end{tabular}
\caption{Performance on TallyQA using accuracy (\%) and RMSE showing the advantage of using background relationships compared to a version of RCN that omits them. \label{tbl:ablation} }
\end{table}

\subsubsection{HowMany-QA.}
HowMany-QA is made by combining counting questions from VQA2 and Visual Genome, so good performance on it serves as a surrogate for good performance on VQA2. HowMany-QA is the best-known dataset for open-ended counting. RCN, IRLC, and Zhang~\etal all use identical region proposals and CNN features.

RCN obtains the highest accuracy on HowMany-QA, outperforming IRLC, which was the best-known result. Zhang~\etal achieves the third-highest accuracy. 
\inline{kim2018bilinear} used the Zhang~\etal method to answer VQA2's counting questions. Although they achieved only third best overall in the CVPR 2018 VQA2 Workshop Challenge, they won for number questions. 

\subsubsection{TallyQA.}
 Example outputs for TallyQA are shown in Fig.~\ref{fig:result-images}.  IRLC's authors were unable to share code with us, so we could not test IRLC on TallyQA. Zhang~\etal uses the same Faster R-CNN region proposals and CNN features as RCN.
 
For Test-Simple, RCN achieves the best accuracy, with Zhang~\etal performing only slightly worse. On Test-Complex, RCN also achieves the highest accuracy. The next best method is again Zhang~\etal, but there is a greater  gap between the two models. This may be because Zhang~\etal does not have an explicit mechanism for relational reasoning between objects and backgrounds, potentially impairing its ability to identify duplicates and compare attributes from different image regions.

Consistent with our claim that complex questions require more than detection, DETECT is the worst performer on Test-Complex. DETECT performs better on Test-Simple, but there is still a large gap between it and RCN.

To study the importance of the object-background model, we ran RCN without the $RN(O,B)$ component. As seen in Table~\ref{tbl:ablation}, this hurts performance for both simple and complex questions showing the value of the background model.

\subsubsection{Positional Reasoning Questions.}
Since RCN uses object-based relational reasoning, we expect it to outperform other methods for positional reasoning questions. To study this, we filtered out positional reasoning questions from TallyQA's Test-Complex set using common qualifiers such as \textit{left, right, top, up, bottom, near, on, in}, and then we measured accuracy for Zhang~\etal and RCN. We found that RCN outperformed Zhang~\etal's model by 6.38\% absolute for these questions, which further demonstrates RCN's  efficacy.

\subsection{Performance Without Location Features}
To assess the impact of using the spatial location information of each proposal, we conducted an experiment in which we removed the location features $s_{ij}$ given to RCN. For HowMany-QA, removing location caused a 5.4\% decrease in accuracy (absolute). For TallyQA, it caused a decrease of 2.8\% accuracy (absolute) for Test-Simple and 2.4\% accuracy (absolute) for Test-Complex.

\subsection{Comparison with the Original RN}
The original RN model uses raw CNN feature maps, rather than region proposals. Running the original RN model on HowMany-QA, it achieved 3.46 RMSE and about 20\% less accuracy than RCN. RCN likely achieves better performance due to its improved architecture and due to using region proposals.

\section{Visualizing RCN}

\begin{figure*}[t!]
\centering
\footnotesize
    \captionsetup[subfigure]{justification=centering}     
        \begin{subfigure}[t]{0.28\textwidth}
		\includegraphics[width=1.1\textwidth, ]{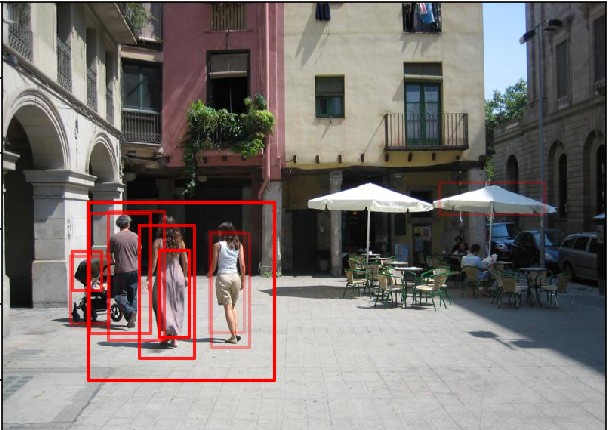}
        \caption{
	    How many people are wearing long dresses? \label{subfig:d} }
    \end{subfigure}
    \hfill    
            \begin{subfigure}[t]{0.28\textwidth}
        	 	\includegraphics[width=1.12\textwidth, ]{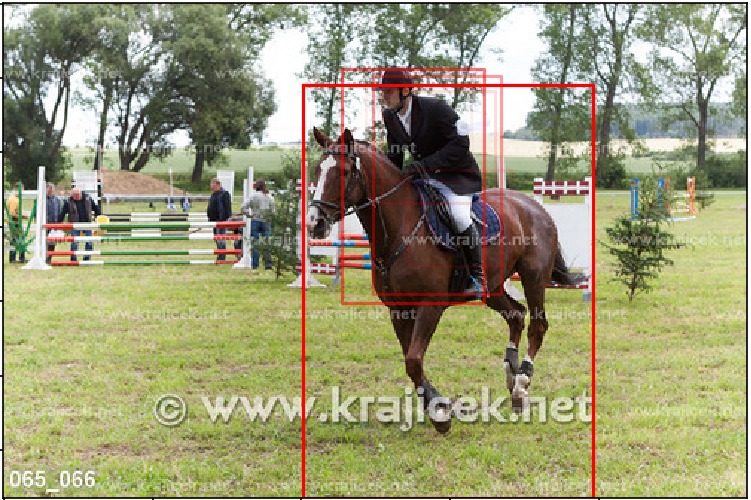}
        \caption{
	    How many people are sitting on a horse? \label{subfig:e}}
    \end{subfigure}   
    \hfill
    \begin{subfigure}[t]{0.28\textwidth}
	 	\includegraphics[width=1.1\textwidth, ]{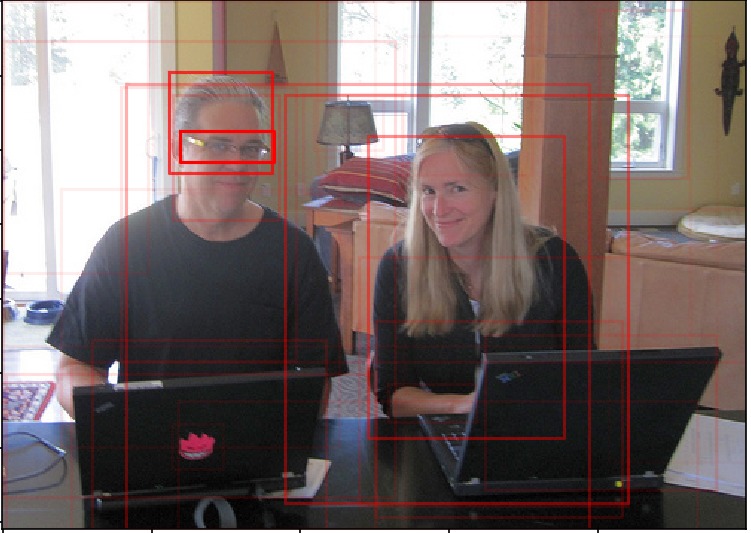}
\caption{How many people are wearing glasses? \label{subfig:c} }
\end{subfigure}
    \begin{subfigure}[t]{0.28\textwidth}
	 	\includegraphics[width=1.1\textwidth, ]{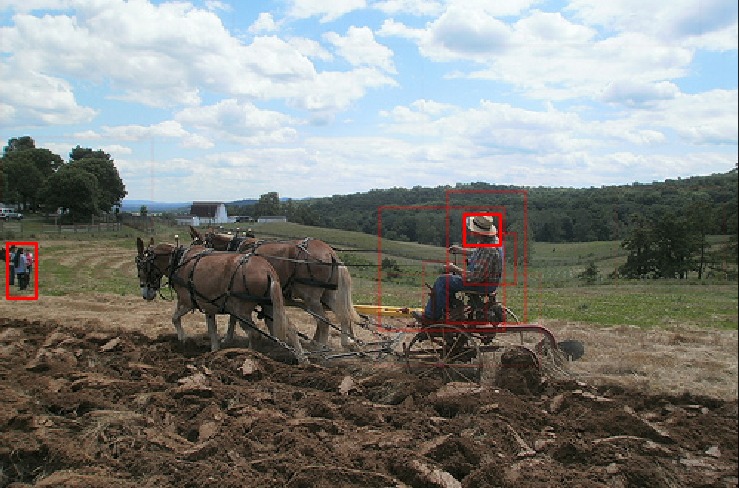}
        \caption{
	    How many people have a hat? \label{subfig:b} }
	\end{subfigure}
    \hfill
    \begin{subfigure}[t]{0.28\textwidth}
		\includegraphics[width=1.1\textwidth, ]{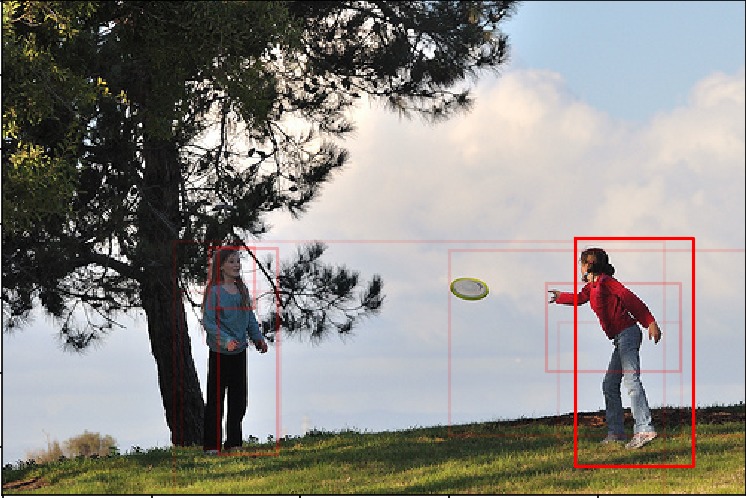}
        \caption{
	    How many players are wearing red? \label{subfig:a}}   
	\end{subfigure}
    \hfill    
    \begin{subfigure}[t]{0.28\textwidth}
	 	\includegraphics[width=1.1\textwidth, ]{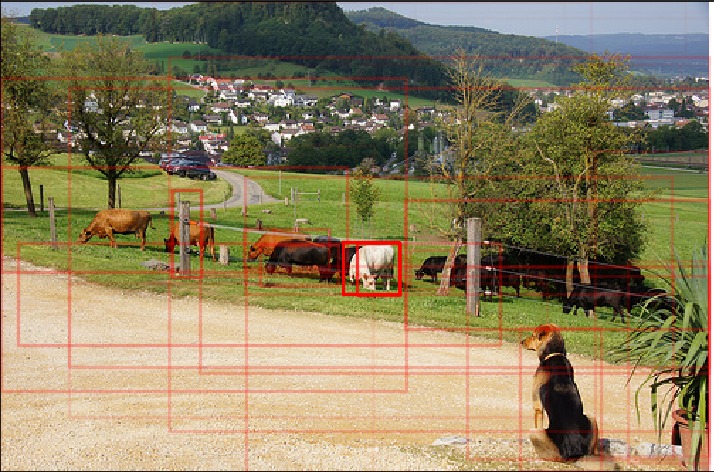}
\caption{How many white cows are there?\label{subfig:f} }
\end{subfigure}   
    \begin{subfigure}[t]{0.28\textwidth}
	 	\includegraphics[width=1.1\textwidth, ]{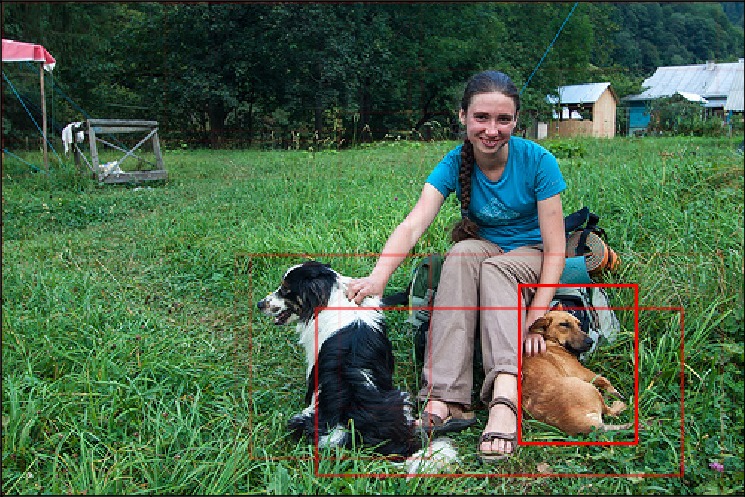}
        \caption{
	    How many dogs are sleeping in the image? \label{subfig:b} }
	\end{subfigure}
    \hfill
    \begin{subfigure}[t]{0.28\textwidth}
		\includegraphics[width=1.1\textwidth, ]{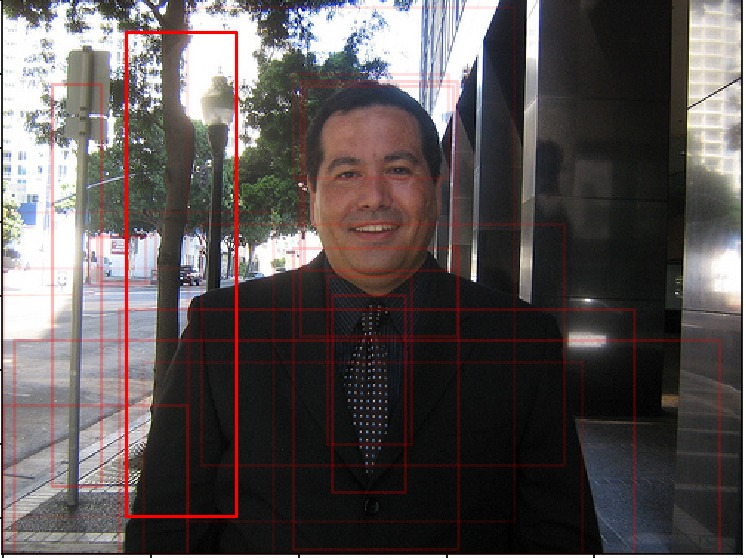}
        \caption{
	    How many street lights are to be seen behind this man? \label{subfig:a}}   
	\end{subfigure}
    \hfill    
    \begin{subfigure}[t]{0.28\textwidth}
	 	\includegraphics[width=1.1\textwidth, ]{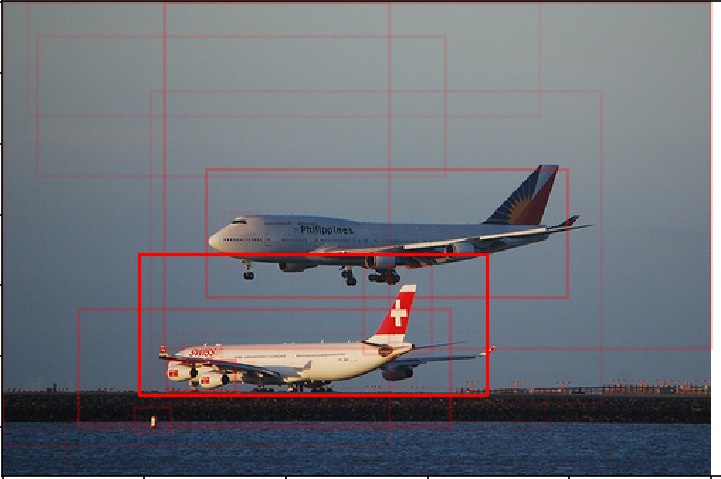}
\caption{How many of the planes are on the ground?\label{subfig:f} }
\end{subfigure}   
\caption{ Modified Grad-CAM visualizations show where RNC  is looking to make predictions. The importance of each object proposals is proportional to the color intensity of the bounding boxes. \label{fig:grad-cam} }
    \end{figure*}

To visualize RCN's inference process, we modified Grad-CAM~\cite{selvaraju2017grad}. Grad-CAM is a technique that, for a given prediction, generates a coarse heat map based on the gradient flow in the final convolutional layers. To adapt Grad-CAM to RCN, it is necessary to derive scores for each proposal. To do this, we first find the pairwise object-background score $score(o_i,b_j)$ using the gradient obtained at layer $g_{\theta_2}$. We then assign a score to each proposal using $ score(o_i) =  \max_{i,j}score(o_i, b_j) $. Scores for all proposals are then scaled from 0 to 1 and visualized on the original image. The results are shown in Fig.~\ref{fig:grad-cam}.

\section{Discussion}
\vspace{5pt}
RCN achieved state-of-the-art results across all of the datasets, even outperforming Zhang~\etal\!, which is the best published result on VQA2's counting questions, and IRLC, which was the previous best result on HowMany-QA. The same regions and visual features were used across RCN, Zhang~\etal, and IRLC, so the difference in performance is not due to using superior visual features, which is a frequent confound in many works.  Our experiments showcased that there is a large performance gap between the ability for models to answer simple and complex questions. This gap was especially large for RCN and the Zhang~\etal method. A likely reason is that more data is required for complex questions to handle the full range of attributes and relations. 

We found that region based methods, such as RCN, IRLC, and the Zhang~\etal  model have better results compared to direct methods, e.g., {MUTAN}~\cite{ben2017mutan}. However, all of these region based models, including ours, are not based on actual \textit{nameable} objects but \textit{object proposals}, which range from 10--100 in number for each image and can consist of many non-object regions and  overlapping boxes. Intelligently pruning/refining of these proposals may improve performance of these systems. We tried simple non-maximal suppression to prune out the overlapping boxes for RCN, but it did not improve performance. We believe this to be due to the relational capacities of RCN which can learn to ignore duplicate or similar boxes based on the features and positions of the boxes more intelligently than off-the-shelf non-maximal suppression.

\section{Conclusions}
In this paper, we distinguished between simple and complex open-ended counting questions in VQA, where simple questions could be correctly answered using object detection alone. To do this, we created TallyQA, the world's largest dataset for open-ended counting using VQA, which will be made publicly available. We also described the RCN framework and showed that it can effectively answer both simple and complex counting questions compared to baseline models and state-of-the-art approaches for open-ended counting. RCN combines region proposals with relationship networks, enabling them to be efficiently used with high-resolution imagery. We found that RCN worked especially well compared to others on complex questions. Our work better defines the issues with open-ended counting, and sets the stage for future work on this problem.

\section*{Acknowledgments} We thank Robik Shrestha for useful discussions. The lab thanks NVIDIA for the donation of a GPU, and we thank Alexander Trott for assistance with HowMany-QA.

\bibliographystyle{aaai}
\bibliography{egbib}

\end{document}